\documentclass[runningheads]{llncs}
\usepackage[T1]{fontenc}
\usepackage{float} 
\usepackage{hyperref}

\usepackage{subcaption}
\usepackage{caption}
\usepackage{graphicx,verbatim}
\usepackage{amsmath}
\usepackage{amssymb}
\usepackage{multirow}
\usepackage[table,xcdraw]{xcolor}
\usepackage{multirow}
\usepackage{booktabs}
\usepackage{algorithm}
\usepackage{algorithmic}
\usepackage{times}
\usepackage{epsfig}
\usepackage{graphicx}
\usepackage{amsmath}

\usepackage{amsthm}
\usepackage{amssymb}
\usepackage{mathrsfs}
\usepackage{multirow}
\usepackage{threeparttable}
\usepackage{xcolor}
\usepackage{color}

\begin{document}
\title{Pose as Clinical Prior: Learning Dual Representations for Scoliosis Screening}
\titlerunning{Learning Dual Representations for Scoliosis Screening}
%
%
\authorrunning{Z. Zhou et al.}  
 \author{
   Zirui Zhou\inst{1} 
   \and Zizhao Peng\inst{1,2} 
   \and Dongyang Jin\inst{1}
   \and Chao Fan\inst{3} 
   \and \\ Fengwei An\inst{1} 
   \and Shiqi Yu\inst{1}$^{*}$ 
 }
\institute{Southern University of Science and Technology, Shenzhen, China 
\and 
The Hong Kong Polytechnic University, Hong Kong, China
\and 
School of Artificial Intelligence, Shenzhen University, Shenzhen, China
}


\maketitle              
\begin{abstract}
Recent AI-based scoliosis screening methods primarily rely on large-scale silhouette datasets, often neglecting clinically relevant postural asymmetries—key indicators in traditional screening. In contrast, pose data provide an intuitive skeletal representation, enhancing clinical interpretability across various medical applications. However, pose-based scoliosis screening remains underexplored due to two main challenges: (1) the scarcity of large-scale, annotated pose datasets; and (2) the discrete and noise-sensitive nature of raw pose coordinates, which hinders the modeling of subtle asymmetries. To address these limitations, we introduce \textbf{Scoliosis1K-Pose}, a 2D human pose annotation set that extends the original Scoliosis1K dataset, comprising 447,900 frames of 2D keypoints from 1,050 adolescents. Building on this dataset, we introduce the \textbf{Dual Representation Framework (DRF)}, which integrates a continuous \textit{skeleton map} to preserve spatial structure with a discrete \textit{Postural Asymmetry Vector (PAV)} that encodes clinically relevant asymmetry descriptors. A novel \textit{PAV-Guided Attention (PGA)} module further uses the PAV as clinical prior to direct feature extraction from the skeleton map, focusing on clinically meaningful asymmetries. Extensive experiments demonstrate that DRF achieves state-of-the-art performance. Visualizations further confirm that the model leverages clinical asymmetry cues to guide feature extraction and promote synergy between its dual representations. The dataset and code are publicly available at \url{https://zhouzi180.github.io/Scoliosis1K/}.


\end{abstract}
\section{Introduction}
Scoliosis is a complex, three-dimensional spinal deformity that affects approximately 3.1\% of adolescents worldwide~\cite{li2024prevalence}. If untreated, it may lead to chronic pain, respiratory problems, and a diminished quality of life~\cite{weinstein2008adolescent}. Early and accurate screening is therefore essential. Traditional screening methods such as visual inspection, the Adam's forward bend test, and trunk rotation measurements are time-consuming, require specialized clinical expertise, and raise privacy concerns due to the need for exposed torso examinations. These limitations hinder large-scale screening, especially in resource-constrained settings~\cite{kadhim2020status}.

Artificial intelligence (AI) offers a promising solution to overcome these limitations. Early AI-based methods relied on static trunk imaging using photogrammetry~\cite{zhang2023deep,yang2019development} or 3D surface topography~\cite{adankon2012non,adankon2013scoliosis} to assess back asymmetry. While reducing the need for expert interpretation, these methods still depend on controlled environments and require exposed torsos. Recent approaches explore dynamic biomarkers in naturalistic settings. Visual gait analysis, among these methods, captures walking-related movements and enhances both privacy and scalability in scoliosis screening. Zhou et al.~\cite{zhou2024gait} introduced this paradigm with the Scoliosis1K dataset, comprising human silhouettes extracted from natural walking videos. They also introduced ScoNet-MT, a CNN-based model designed to extract fine-grained body shape features for scoliosis screening. Despite improving data accessibility and reducing acquisition constraints, these methods often overlook clinically relevant asymmetries—such as shoulder imbalance and pelvic tilt~\cite{cma2020guideline}—that are critical for accurate and interpretable assessments.

Human pose data, with its explicit skeletal structure, offers a natural way to integrate these clinical evaluation criteria and has demonstrated value in various medical applications~\cite{lu2020vision,quan2024causality,islam2023using,wang2024enhancing}. Despite its potential, the use of pose data in scoliosis screening remains underexplored, mainly due to two key challenges: (1) the scarcity of large-scale, annotated pose datasets; and (2) the discrete and noise-sensitive nature of raw pose coordinates, which hinders the modeling of subtle body asymmetries.

To bridge this gap, we introduce \textbf{Scoliosis1K-Pose}, a 2D human pose annotation set that augments the original Scoliosis1K dataset~\cite{zhou2024gait} for scoliosis screening. It provides 2D keypoints extracted from natural walking videos of adolescents.	Each frame includes 17 keypoints in the MS-COCO format~\cite{coco}, extracted using ViTPose~\cite{vitpose}. Building on this dataset, we propose the \textbf{Dual Representation Framework (DRF)}, a pose-based method for scoliosis screening. DRF integrates two complementary pose representations: a continuous \emph{skeleton map} that preserves the global spatial structure of the pose, and a discrete \emph{Postural Asymmetry Vector (PAV)} that explicitly encodes  clinically relevant asymmetry descriptors (e.g., vertical, midline, and angular deviations) between paired keypoints. To synergize these representations, our novel \emph{PAV-Guided Attention (PGA)} employs the PAV as a clinical prior to guide the network’s feature extraction, directing attention to the most clinically relevant asymmetries within the skeleton map. This enables DRF to capture subtle yet clinically meaningful postural patterns for scoliosis assessment.

Our main contributions are threefold:

\textbf{- Scoliosis1K-Pose:} A new 2D human pose annotation set introduced in this work, augmenting the original Scoliosis1K dataset with detailed keypoint annotations for scoliosis screening.

\textbf{- Dual Representation Framework (DRF):} A novel, clinically-informed framework that synergistically combines continuous skeleton maps with discrete, clinically-derived PAV. This is implemented through a PGA module that embeds clinical prior into the feature learning process.

\textbf{- State-of-the-Art Performance:} Extensive experiments demonstrate that DRF can achieve superior performance in scoliosis screening. Visualizations further confirm that the model leverages clinical asymmetry cues to guide feature extraction and promote synergy between its dual representations.

\begin{figure*} [!t]
        \centering
	\includegraphics[width=0.7\textwidth]{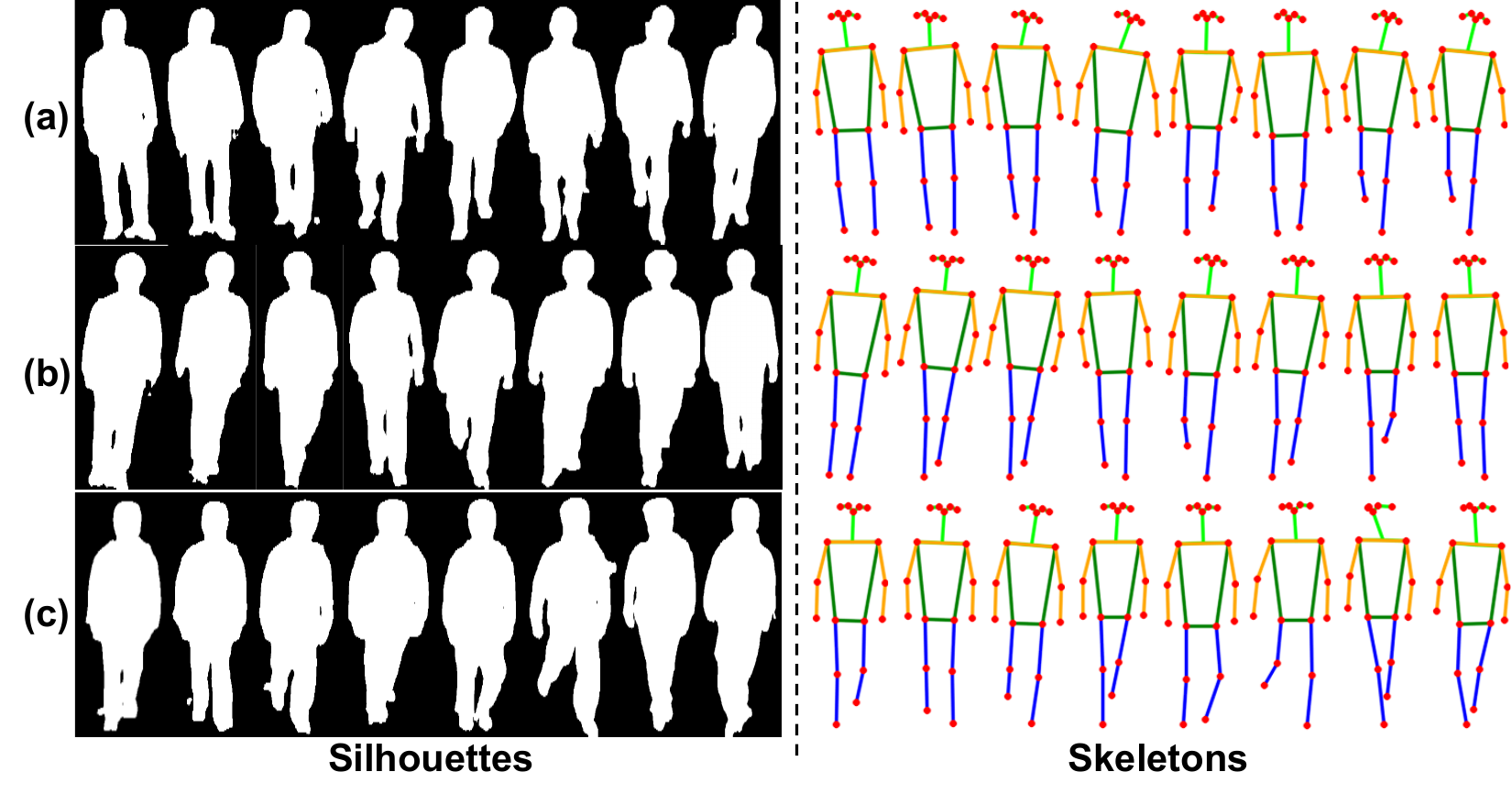}
\caption{Examples from the Scoliosis1K dataset. Left: silhouettes representing (a) positive, (b) neutral, and (c) negative cases. Right: corresponding 2D skeletal keypoints estimated by ViTPose~\cite{vitpose}.}
	\label{fig:dataset}
\end{figure*}

\section{Dataset}\label{sec:dataset}
Our research is built upon the Scoliosis1K dataset~\cite{zhou2024gait}, which originally provided only silhouette data from 1,493 walking videos of 1,050 adolescents. To enable fine-grained postural analysis, we introduce \textit{Scoliosis1K-Pose}, a large-scale 2D human pose annotation set that extends the original Scoliosis1K.

Scoliosis1K-Pose was constructed by processing 447,900 frames using ViTPose~\cite{vitpose} to extract 17 anatomical keypoints per frame, following the MS-COCO format~\cite{coco}. The extracted keypoints provide detailed skeletal representations of facial features (eyes, ears, nose), upper limbs (shoulders, elbows, wrists), and lower limbs (hips, knees, ankles).	This augmentation transforms Scoliosis1K into a comprehensive multi-modal resource that integrates silhouette and 2D pose data, as illustrated in Figure~\ref{fig:dataset}. Participants are categorized as positive (scoliosis), neutral (borderline), or negative (non-scoliosis) according to established screening protocols. For detailed information on the original data collection and its characteristics, please refer to~\cite{zhou2024gait}.

\begin{figure*} [t]
        \centering
	\includegraphics[width=0.75\textwidth]{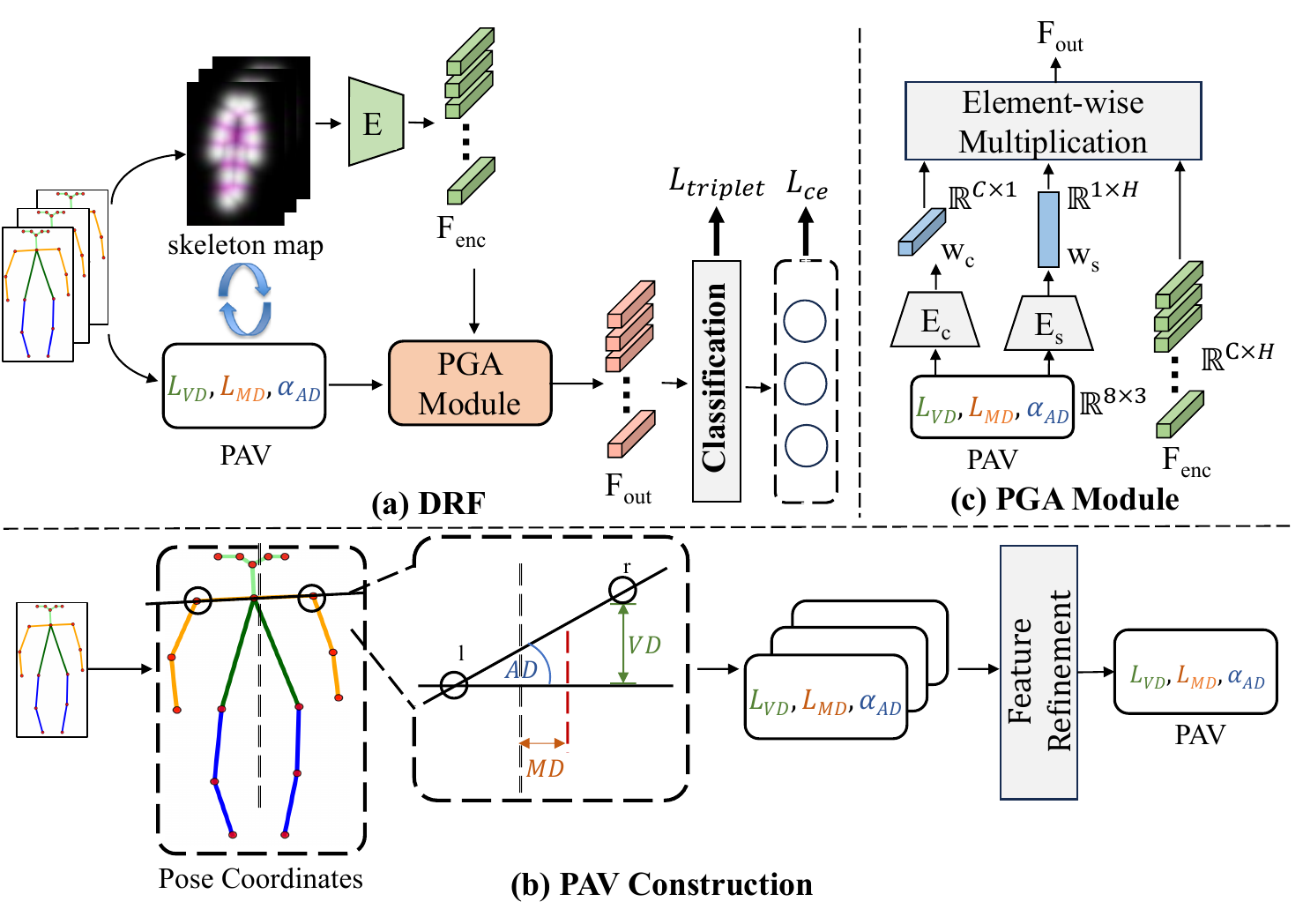}
\caption{Overview of the proposed Dual Representation Framework (DRF). (a) The overall pipeline, from raw pose data to the final scoliosis assessment. (b) The construction process of the Postural Asymmetry Vector (PAV). (c) The architecture of the PAV-Guided Attention (PGA) module, which generates channel and spatial attention weights.}
\label{fig:drf}
\end{figure*}

\section{Method}\label{sec:method}
We propose the \textbf{Dual Representation Framework (DRF)}, a novel approach to scoliosis screening that explicitly integrates clinical priors into a deep learning pipeline. The core idea is to emulate how clinicians assess scoliosis by identifying postural asymmetries. As shown in Figure~\ref{fig:drf}(a), the framework first transforms raw pose data into two complementary representations: a continuous \emph{skeleton map} and a discrete \emph{Postural Asymmetry Vector (PAV)}. The skeleton map is processed by a feature encoder.	Concurrently, the PAV, serving as a clinical prior, guides a \emph{PAV-Guided Attention (PGA)} module to generate attention weights that modulate the extracted features. Finally, the refined features are used for scoliosis assessment.	

The following sections describe the construction of the dual pose representations (Section~\ref{dual_rep}) and the PAV-guided feature learning and assessment (Section~\ref{cpi}).	

\subsection{Dual Pose Representations}\label{dual_rep}
Given a sequence of 2D pose keypoints, we first normalize the data to ensure robustness to variations in subject position and scale. This involves aligning the pelvis by translating the midpoint of the hip joints to the origin, followed by height normalization to a standard height of 128 pixels. From the normalized pose, we construct two complementary representations.		

\paragraph{\textbf{Skeleton Map: Continuous Gaussian Heatmaps.}}\label{skeleton_maps}
To convert sparse 2D keypoints into a dense, silhouette-like representation suitable for convolutional networks, we construct a two-channel skeleton map. This approach, inspired by related works~\cite{fan2024skeletongait,duan2022revisiting}, provides a richer input than raw coordinates by capturing both local joint details and global skeletal structure. It consists of two channels:

\textbf{- Keypoint Map:} Encodes each joint's location and confidence with a Gaussian heatmap:	
\begin{equation}
    J(i,j) = \sum_{k} \exp\left(-\frac{(i-x_k)^2+(j-y_k)^2}{2\sigma^2}\right) \cdot c_k,
\end{equation}
where $(x_k, y_k)$ and $\sigma$ are the coordinates and confidence score of the $k$-th joint, respectively.

\textbf{- Limb Map:} Represents skeletal connections as line-like heatmaps, preserving structural integrity:
\begin{equation}
    L(i,j) = \sum_{n} \exp\left(-\frac{d_{L2}((i,j), S[n^-, n^+])^2}{2\sigma^2}\right) \cdot \min(c_{n^-}, c_{n^+}),
\end{equation}
where $S[n^-, n^+]$ denotes the limb segment between joints $n^-$ and $n^+$, and $d_{L2}(\cdot)$ is the Euclidean distance from pixel $(i,j)$ to this segment.

The resulting skeleton map yields a robust, continuous representation that mitigates pose estimation noise and serves as the primary input for feature extraction.	

\paragraph{\textbf{PAV: Discrete Clinical Prior.}}\label{pav}
To explicitly incorporate clinical priors into our framework, we introduce the Postural Asymmetry Vector (PAV). Instead of relying on a few hand-picked landmarks, which are susceptible to noise and may miss subtle compensatory patterns, the PAV is designed to capture a holistic asymmetry profile. The PAV is a matrix $v \in \mathbb{R}^{P \times M}$, where $P$ denotes the number of anatomically symmetric keypoint pairs (e.g., eyes, ears, shoulders, elbows, wrists, hips, knees, and ankles in COCO17 format), with the $p$-th pair represented as $(k_p^L, k_p^R)$, and $M$ denotes the number of asymmetry metrics.	

The PAV is constructed in two stages, as illustrated in Fig.~\ref{fig:drf}(b):	

\textbf{1. Asymmetry Metric Computation.} For each frame, a raw matrix $\tilde{v} \in \mathbb{R}^{P \times M}$ is computed using the following three metrics for each keypoint pair:	

- Vertical Deviation ($L_{VD}$): $|y_p^L - y_p^R|$, quantifies height differences, directly corresponding to clinical signs like shoulder and pelvic imbalance.

- Midline Deviation ($L_{MD}$): $\left|\frac{x_p^L + x_p^R}{2} - x_{\text{midline}}\right|$, measures lateral drift from the body's central axis ($x_{\text{midline}}$ is defined by the hip center), reflecting spinal curvature.

- Angular Deviation ($\alpha_{AD}$): $\left|\arctan\left(\frac{y_p^L - y_p^R}{x_p^L - x_p^R}\right)\right|$, captures the tilt of the segment connecting a keypoint pair, indicative of rotational asymmetries in the trunk and limbs.

\textbf{2. Sequence-Level PAV Refinement.} The raw matrix $\tilde{v}$ is refined across the sequence to obtain a stable PAV $v$ using the following steps:	

(1) Outlier Removal: Statistical filtering is applied independently to each element in $\tilde{v}$ using the interquartile range (IQR) method.

(2) Temporal Aggregation: For each metric, the mean of valid measurements across all frames is computed, summarizing temporal variation into a stable sequence-level descriptor.

(3) Normalization: Min-Max scaling is applied across the dataset for all $P \times M$ dimensions, standardizing values to the $[0,1]$ range and ensuring consistent feature scaling.

This two-stage design enables robust quantification of global postural asymmetry and allows the model to capture complex, clinically relevant inter-joint relationships. The resulting PAV acts as a clinical prior to guide the subsequent feature learning process.

\subsection{PAV-Guided Feature Learning and Assessment}\label{cpi}
Our DRF uses the clinical prior encoded in the PAV to guide visual feature learning from the skeleton map. This is accomplished using a feature encoder followed by our novel PAV-Guided Attention (PGA) module.

\paragraph{\textbf{Feature Encoder.}}\label{feature_encoder} 
First, the feature encoder processes the input skeleton map sequence to extract high-level features. For a fair comparison, we adopt the encoder architecture from ScoNet-MT~\cite{zhou2024gait}.	It uses a ResNet-based backbone followed by temporal and horizontal pooling to generate a feature map $F_{\text{enc}} \in \mathbb{R}^{C \times H}$, where $C=256$ is the number of channels and $H$ is the number of horizontal body segments. This feature map $F_{\text{enc}}$ captures general postural patterns from the data.		

\paragraph{\textbf{Prior-Guided Attention (PGA) Module.}}\label{pav_attention}
To ensure these features are clinically relevant, we introduce the PGA Module. Unlike standard attention that relies solely on data-driven correlations, our module explicitly incorporates a clinical prior to generate attention weights. In our framework, the prior is the PAV, which guides the model to focus on features critical for scoliosis screening.

As shown in Fig.~\ref{fig:drf}(c), the module receives two inputs: the encoded features $F_{\text{enc}}$ and the PAV $v$. It then generates both channel-wise ($w_c$) and spatial-wise ($w_s$) attention weights using the PAV:
\[
w_c = \sigma(W_c \cdot v), \quad w_s = \sigma(W_s * v),
\]
where $\sigma$ denotes the sigmoid function, $W_c$ is a learnable linear layer, and $W_s$ is a learnable 1D convolution. The resulting attention weights, $w_c \in \mathbb{R}^{C \times 1}$ and $w_s \in \mathbb{R}^{1 \times H}$, are used to recalibrate the encoded features via element-wise multiplication:	
\[
F_{\text{out}} = F_{\text{enc}} \odot w_c \odot w_s,
\]
where $\odot$ denotes element-wise multiplication with broadcasting. This produces a refined feature map $F_{\text{out}}$ that integrates visual patterns with clinical priors, yielding a more informative representation for downstream assessment.

\paragraph{\textbf{Scoliosis Assessment and Training Objective.}}
The refined features $F_{\text{out}}$ are passed to a classification head with fully connected layers to produce the final logits for scoliosis assessment. The entire framework is trained end-to-end with a combined loss function, following the approach of ScoNet-MT~\cite{zhou2024gait}:	
\[
L_{total} = L_{ce} + L_{triplet}.
\]

\begin{table}[t] 
\centering
\begin{minipage}[!h]{0.58\textwidth}
\caption{Comparison with state-of-the-arts on Scoliosis1K dataset. The best results are in bold. ScoNet-MT$^{ske}$ refers to ScoNet-MT~\cite{zhou2024gait} adapted for skeleton map input.}
    \label{tab:main_results}
    \centering
    \begin{threeparttable}
    \resizebox{0.94\textwidth}{!}{
\begin{tabular}{c|c|c|ccc}
    \toprule
 &
   &
   &
  \multicolumn{3}{c}{Macro-average} \\ \cline{4-6} 
\multirow{-2}{*}{Input} &
  \multirow{-2}{*}{Method} &
  \multirow{-2}{*}{\begin{tabular}[c]{@{}c@{}}Total\\ Accuracy\end{tabular}} &
  Prec &
  Rec &
  F1 \\ \hline
Silhouette &
  ScoNet-MT~\cite{zhou2024gait} &
  82.0 &
  84.0 &
  75.9 &
  75.4 \\ \hline
 &
  OF-DDNet~\cite{lu2020vision} &
  74.1 &
  68.6 &
  68.4 &
  67.0 \\
\multirow{-2}{*}{\begin{tabular}[c]{@{}c@{}}Pose\\ Coordinates\end{tabular}} &
  GPGait~\cite{fu2023gpgait} &
  74.9 &
  69.2 &
  68.9 &
  67.5 \\ \hline
 &
  ScoNet-MT$^{ske}$ &
  82.5 &
  81.4 &
  74.3 &
  76.6 \\
\multirow{-2}{*}{Skeleton Map} &
  \cellcolor[HTML]{E5E5E5} DRF (ours) &
  \cellcolor[HTML]{E5E5E5}\textbf{86.0} &
  \cellcolor[HTML]{E5E5E5}\textbf{84.1} &
  \cellcolor[HTML]{E5E5E5}\textbf{79.2} &
  \cellcolor[HTML]{E5E5E5}\textbf{80.8} \\ \hline
\end{tabular}}
    \end{threeparttable}
\end{minipage}%
\begin{minipage}[!h]{0.40\textwidth}
    \centering
    \includegraphics[width=0.8\textwidth]{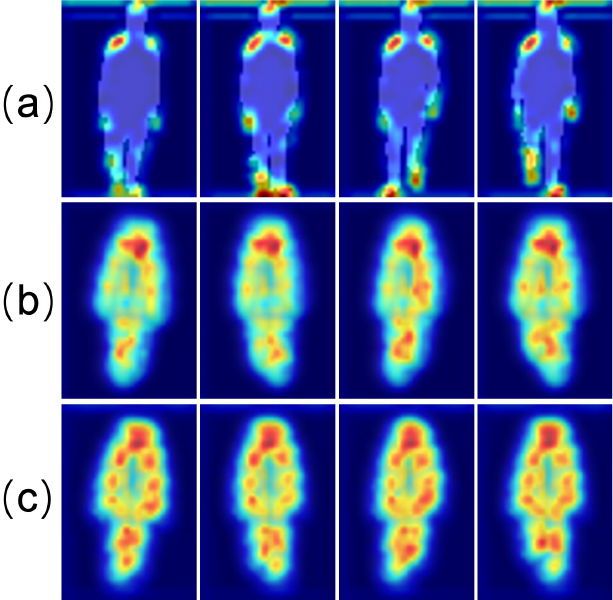}
    \captionsetup{type=figure}
      \caption{Visualization of feature response heatmaps. (a) ScoNet-MT. (b) ScoNet-MT$^{ske}$. (c) Our DRF.}
    \label{fig:heatmap}
\end{minipage}
\end{table}

\section{Experiments}

\subsection{Setup}
\paragraph{\textbf{Evaluation Protocol.}} We use the standard dataset split from~\cite{zhou2024gait} (745 training, 748 test sequences) with real-world class distribution (positive:neutral:negative = 1:1:8). To address class imbalance and enable fair comparison, we report overall accuracy along with macro-averaged precision, recall, and F1-score.	

\paragraph{\textbf{Implementation Details.}} Our method is implemented using PyTorch~\cite{paszke2019pytorch} and OpenGait~\cite{opengait}. Following ScoNet-MT~\cite{zhou2024gait}, we retain the same network architecture and training protocol, with two key modifications: (1) the Conv0 layer is adapted to process skeleton maps instead of silhouettes, and (2) the proposed PGA module is integrated to guide feature extraction.

\subsection{Comparison with State-of-the-Art Methods}
We perform comprehensive comparisons with state-of-the-art methods using various input representations. For silhouette-based methods, we compare our model with ScoNet-MT~\cite{zhou2024gait}.	For pose coordinate-based approaches, we include OF-DDNet~\cite{lu2020vision} and GPGait~\cite{fu2023gpgait}.	OF-DDNet is adapted to process 2D pose inputs to ensure fair comparison.

\paragraph{\textbf{Quantitative Results.}} As shown in Table~\ref{tab:main_results}, our proposed DRF achieves state-of-the-art performance across all evaluation metrics. Pose coordinate-based methods, such as GPGait and OF-DDNet, yield the lowest performance, with F1-scores of 67.5\% and 67.0\%, respectively. Image-based representations perform better overall. The silhouette-based ScoNet-MT model significantly outperforms pose coordinate-based methods. When adapted to use our skeleton map input (ScoNet-MT$^{ske}$), the model achieves an additional 0.5\% improvement in accuracy and a 1.2\% gain in F1-score. These results confirm that the skeleton map captures rich structural information. Importantly, integrating the PGA module with PAV enables our final DRF model to significantly outperform the ScoNet-MT$^{ske}$ baseline. Specifically, accuracy increases by 3.5\%, and the F1-score improves by 4.2\%. This significant improvement demonstrates that explicitly guiding the model with clinical priors on postural asymmetry is highly effective for scoliosis screening.	

\paragraph{\textbf{Qualitative Results.}} As shown in 
Figure~\ref{fig:heatmap}, we visualize the intermediate feature responses using the visualization technique from~\cite{zhou2016learning}. ScoNet-MT with silhouette input (Fig.\ref{fig:heatmap} (a)) focuses on isolated regions (e.g., head and shoulders), missing holistic structural relationships crucial for scoliosis screening. With skeleton map as input,  ScoNet-MT$^{ske}$ (Fig.\ref{fig:heatmap} (b)) preserves structural information but mainly attends to one body side. By incorporating our PGA module guided by the PAV, our DRF (Fig.~\ref{fig:heatmap} (c)) more effectively attends to key regions on both sides of the body (e.g., imbalanced shoulders and pelvic), which strongly corresponds to our clinical prior integration objective. These visualizations confirm that our PGA module, when guided by PAV, effectively guides feature extraction using clinical asymmetry cues, creating the synergistic relationship between our dual representations.

\begin{table*}[t]
    \caption{Ablation studies on Scoliosis1K dataset. The best results are in bold.}
    \centering
    \begin{minipage}{0.52\linewidth}
        \centering
        \resizebox{0.7\linewidth}{!}{
\begin{tabular}{cc|cccc}
\hline
             &              &      & \multicolumn{3}{c}{Macro-average} \\ \cline{4-6} 
\multirow{-2}{*}{Channel} & \multirow{-2}{*}{Spatial} & \multirow{-2}{*}{\begin{tabular}[c]{@{}c@{}}Total\\ Acc\end{tabular}} & Prec          & Rec  & F1            \\ \hline
             &              & 82.5 & 81.4   & 74.3            & 76.6   \\
$\checkmark$ &              & 79.7 & 80.9   & 75.7            & 74.9   \\
             & $\checkmark$ & 81.7 & 77.1   & \textbf{80.8}   & 78.1   \\
 
$\checkmark$              & $\checkmark$              & \textbf{86.0}                                                         & \textbf{84.1} & 79.2 & \textbf{80.8} \\ \hline
\end{tabular}
        }
        \subcaption{Component Analysis of PGA Module. All variants of the PGA module are guided by our PAV.}
        \label{tab:pga_ablation}
    \end{minipage}
    \hfill
    \begin{minipage}{0.44\linewidth}
        \centering
        \resizebox{0.8\linewidth}{!}{
\begin{tabular}{c|cccc}
\hline
\multirow{2}{*}{Guidance Source} & \multirow{2}{*}{\begin{tabular}[c]{@{}c@{}}Total\\ Acc\end{tabular}} & \multicolumn{3}{c}{Macro-average} \\ \cline{3-5} 
               &               & Prec          & Rec           & F1            \\ \hline
Self-attention & 81.4          & \textbf{84.5} & 73.9          & 75.3          \\ \hline
All-Ones       & 82.2          & 80.8          & 74.3          & 74.7          \\
Random         & 75.4          & 82.2          & 62.7          & 57.9          \\
Learnable      & 69.4          & 84.2          & 53.2          & 56.4          \\
PAV (ours)     & \textbf{86.0} & 84.1          & \textbf{79.2} & \textbf{80.8} \\ \hline
\end{tabular}
        }
        \subcaption{Comparison of PAV and alternative guidance sources for the PGA Module.}
        \label{tab:pav_ablation}
    \end{minipage}
\end{table*}
\subsection{Ablation Studies}\label{sec:ablation}
To validate the design of our framework, we conduct two sets of ablation studies. First, we analyze the effectiveness of the PGA module and its components. Second, we verify that the proposed PAV provides superior guidance compared to alternative strategies.

\paragraph{\textbf{Effectiveness of the PGA Module.}} We first evaluate the overall contribution of the PGA module, followed by an analysis of its channel-wise and spatial-wise attention branches. We use ScoNet-MT$^{ske}$ as our baseline, which operates without any attention guidance.	As shown in Table \ref{tab:pga_ablation}, integrating the full PGA module (guided by PAV) into the baseline significantly improves performance, with a 3.5\% increase in accuracy and a 4.2\% gain in F1-score. Furthermore, both attention branches are essential.	Disabling either the channel or spatial branch results in a notable performance decline compared to the full model. This highlights the synergistic interaction between channel and spatial attention, which is essential for effectively leveraging the clinical cues provided by the PAV.			

\paragraph{\textbf{Effectiveness of PAV as a Clinical Prior.}} 
After confirming the effectiveness of the PGA module’s architecture, we next examine the importance of the guidance source.	We compare the PAV with several alternative guidance vectors within the same PGA module architecture. The evaluated alternatives include:	

(1) \textit{No External Prior (Self-attention)}: Guidance is generated directly from feature maps, following a standard self-attention mechanism. This setup evaluates whether the performance gain arises solely from the attention mechanism rather than from clinical prior.		

(2) \textit{Fixed, Uninformative Priors}: Fixed vectors with the same dimensionality as the PAV are used: (a) All-Ones, representing uniform attention, and (b) Random, representing arbitrary, unstructured guidance. These variants test whether a static structural bias could achieve similar benefits.

(3) \textit{Learnable Prior}: The PAV is replaced by a learnable parameter vector, enabling the network to derive a data-driven guidance signal without explicit clinical constraints.

As shown in Table \ref{tab:pav_ablation}, using PAV to guide the PGA module significantly outperforms all alternatives. The self-attention approach results in a 5.5\% drop in F1-score, indicating that the explicit prior offers greater value than internally derived correlations. The uninformative priors (All-Ones, Random) and the learnable prior also perform significantly worse, with F1-score reductions ranging from 6.1\% to 22.9\%. These results confirm that the performance improvement arises not merely from using an attention module, but crucially from the rich, clinically relevant information encoded in the PAV. This underscores the value of incorporating explicit clinical prior for robust scoliosis screening.				

\section{Conclusion}
This study proposes a novel, clinically informed approach for pose-based scoliosis screening. The proposed Dual Representation Framework (DRF), combined with the Scoliosis1K-Pose dataset, integrates continuous skeletal structures and clinically informed asymmetry descriptors to improve model accuracy and interpretability, enabling accessible and privacy-preserving scoliosis screening. We hope that the work can inspire more research on scoliosis screening using massive video data.

    

\begin{credits}
\subsubsection{\ackname} This work was supported by the National Natural Science Foundation of China (Grant 62476120) and the Scientific Foundation for Youth Scholars of Shenzhen University (Grant 868-000001033383).
\subsubsection{\discintname}
The authors have no competing interests to declare that are relevant to the content of this article.

\end{credits}

%
%
%
%
\bibliographystyle{splncs04}
\bibliography{refs}

\begin{thebibliography}{10}
\providecommand{\url}[1]{\texttt{#1}}
\providecommand{\urlprefix}{URL }
\providecommand{\doi}[1]{https://doi.org/#1}

\bibitem{adankon2013scoliosis}
Adankon, M.M., Chihab, N., Dansereau, J., Labelle, H., Cheriet, F.: Scoliosis follow-up using noninvasive trunk surface acquisition. IEEE Transactions on Biomedical Engineering  \textbf{60}(8),  2262--2270 (2013)

\bibitem{adankon2012non}
Adankon, M.M., Dansereau, J., Labelle, H., Cheriet, F.: Non invasive classification system of scoliosis curve types using least-squares support vector machines. Artificial intelligence in medicine  \textbf{56}(2),  99--107 (2012)

\bibitem{cma2020guideline}
of~the Chinese Orthopaedic~Association, S.S.G.: Clinical practice guideline and pathway for screening adolescent idiopathic scoliosis in china. Chinese Journal of Orthopaedics  \textbf{40}(23),  1574--1582 (2020)

\bibitem{duan2022revisiting}
Duan, H., Zhao, Y., Chen, K., Lin, D., Dai, B.: Revisiting skeleton-based action recognition. In: Proceedings of the IEEE/CVF conference on computer vision and pattern recognition. pp. 2969--2978 (2022)

\bibitem{opengait}
Fan, C., Liang, J., Shen, C., Hou, S., Huang, Y., Yu, S.: Opengait: Revisiting gait recognition towards better practicality. In: Proceedings of the IEEE/CVF Conference on Computer Vision and Pattern Recognition. pp. 9707--9716 (2023)

\bibitem{fan2024skeletongait}
Fan, C., Ma, J., Jin, D., Shen, C., Yu, S.: Skeletongait: Gait recognition using skeleton maps. In: Proceedings of the AAAI conference on artificial intelligence. vol.~38, pp. 1662--1669 (2024)

\bibitem{fu2023gpgait}
Fu, Y., Meng, S., Hou, S., Hu, X., Huang, Y.: Gpgait: Generalized pose-based gait recognition. In: Proceedings of the IEEE/CVF International Conference on Computer Vision. pp. 19595--19604 (2023)

\bibitem{islam2023using}
Islam, M.S., Rahman, W., Abdelkader, A., Lee, S., Yang, P.T., Purks, J.L., Adams, J.L., Schneider, R.B., Dorsey, E.R., Hoque, E.: Using ai to measure parkinson’s disease severity at home. NPJ digital medicine  \textbf{6}(1), ~156 (2023)

\bibitem{kadhim2020status}
Kadhim, M., Lucak, T., Schexnayder, S.: Current status of scoliosis school screening: targeted screening of underserved populations may be the solution. Public Health  \textbf{178},  72--77 (2020)

\bibitem{li2024prevalence}
Li, M., Nie, Q., Liu, J., Jiang, Z.: Prevalence of scoliosis in children and adolescents: a systematic review and meta-analysis. Front Pediatr  \textbf{12},  1399049 (2024)

\bibitem{coco}
Lin, T.Y., Maire, M., Belongie, S., Hays, J., Perona, P., Ramanan, D., Doll{\'a}r, P., Zitnick, C.L.: Microsoft coco: Common objects in context. In: Computer vision--ECCV 2014: 13th European conference, zurich, Switzerland, September 6-12, 2014, proceedings, part v 13. pp. 740--755. Springer (2014)

\bibitem{lu2020vision}
Lu, M., Poston, K., Pfefferbaum, A., Sullivan, E.V., Fei-Fei, L., Pohl, K.M., Niebles, J.C., Adeli, E.: Vision-based estimation of mds-updrs gait scores for assessing parkinson’s disease motor severity. In: Medical Image Computing and Computer Assisted Intervention--MICCAI 2020: 23rd International Conference, Lima, Peru, October 4--8, 2020, Proceedings, Part III 23. pp. 637--647. Springer (2020)

\bibitem{paszke2019pytorch}
Paszke, A., Gross, S., Massa, F., Lerer, A., Bradbury, J., Chanan, G., Killeen, T., Lin, Z., Gimelshein, N., Antiga, L., et~al.: Pytorch: An imperative style, high-performance deep learning library. Advances in neural information processing systems  \textbf{32} (2019)

\bibitem{quan2024causality}
Quan, Y., Zhang, C., Guo, R., Qian, X.: Causality-informed fusion network for automated assessment of parkinsonian body bradykinesia. In: International Conference on Medical Image Computing and Computer-Assisted Intervention. pp. 78--88. Springer (2024)

\bibitem{wang2024enhancing}
Wang, D., Yuan, K., Muller, C., Blanc, F., Padoy, N., Seo, H.: Enhancing gait video analysis in neurodegenerative diseases by knowledge augmentation in vision language model. In: International Conference on Medical Image Computing and Computer-Assisted Intervention. pp. 251--261. Springer (2024)

\bibitem{weinstein2008adolescent}
Weinstein, S.L., Dolan, L.A., Cheng, J.C., et~al.: Adolescent idiopathic scoliosis. Lancet  \textbf{371}(9623),  1527--1537 (2008)

\bibitem{vitpose}
Xu, Y., Zhang, J., Zhang, Q., Tao, D.: Vitpose: Simple vision transformer baselines for human pose estimation. Advances in Neural Information Processing Systems  \textbf{35},  38571--38584 (2022)

\bibitem{yang2019development}
Yang, J., Zhang, K., Fan, H., Huang, Z., Xiang, Y., Yang, J., He, L., Zhang, L., Yang, Y., Li, R., et~al.: Development and validation of deep learning algorithms for scoliosis screening using back images. Communications biology  \textbf{2}(1), ~390 (2019)

\bibitem{zhang2023deep}
Zhang, T., Zhu, C., Zhao, Y., Zhao, M., Wang, Z., Song, R., Meng, N., Sial, A., Diwan, A., Liu, J., et~al.: Deep learning model to classify and monitor idiopathic scoliosis in adolescents using a single smartphone photograph. JAMA Network Open  \textbf{6}(8),  e2330617--e2330617 (2023)

\bibitem{zhou2016learning}
Zhou, B., Khosla, A., Lapedriza, A., Oliva, A., Torralba, A.: Learning deep features for discriminative localization. In: Proceedings of the IEEE conference on computer vision and pattern recognition. pp. 2921--2929 (2016)

\bibitem{zhou2024gait}
Zhou, Z., Liang, J., Peng, Z., Fan, C., An, F., Yu, S.: Gait patterns as biomarkers: A video-based approach for classifying scoliosis. In: International Conference on Medical Image Computing and Computer-Assisted Intervention. pp. 284--294. Springer (2024)

\end{thebibliography}
\end{document}